\documentclass[5p,twocolumn]{elsarticle}

\usepackage{helvet} % Add helvet package
\usepackage{amsmath}            
\usepackage{epstopdf}           
\usepackage{flushend}    % Dunno why       
\usepackage{overpic} % adding annotations over images
\usepackage{tikz}

\usepackage[inkscapelatex=false]{svg}

\usepackage{caption}
\usepackage{subcaption}
\usepackage{xcolor}

\usepackage{graphicx}
\usepackage{pgffor}

\usepackage{xcolor}		% colors for hyperlinks
\usepackage{hyperref}
\hypersetup{
    colorlinks = true,
    linkcolor = magenta,
    urlcolor  = blue,
    citecolor = purple,
    anchorcolor = black,
}   

\usepackage[superscript,biblabel]{cite}
\usepackage{listings}
\usepackage{multirow}
\usepackage[normalem]{ulem}

\usepackage{amssymb}
\usepackage{refcount}
\usepackage{ifthen}
\newcommand{\linerange}[2]{%
\ifthenelse{\equal{\getrefnumber{#1}}{\getrefnumber{#2}}}{%
line \ref{#1}%
}{%
lines \ref{#1}--\ref{#2}%
}%
}

\usepackage[switch]{lineno} % left+right hand side
\biboptions{numbers,square,sort&compress}
\usepackage[figuresright]{rotating}

\newcommand\extrafootertext[1]{%
    \bgroup
    \renewcommand\thefootnote{\fnsymbol{footnote}}%
    \renewcommand\thempfootnote{\fnsymbol{mpfootnote}}%
    \footnotetext[0]{#1}%
    \egroup
}

\newcommand{\labeledtikzfig}[4]{
    \begin{tikzpicture}
        \node[anchor=south west,inner sep=0] (image) at (0,0) {\includegraphics[width=\linewidth]{#1}};
        \begin{scope}[x={(image.south east)},y={(image.north west)}]
            \node [anchor=north west] at (#3,#4) {\color{black}\textbf{(#2)}};
        \end{scope}
    \end{tikzpicture}
}

\begin{document}
\begin{frontmatter}
%%%%%%%%%%%%%%%%   Title   %%%%%%%%%%%%%%%%
\title{
RACER: An LLM-powered Methodology for Scalable Analysis of \\
Semi-structured Mental Health Interviews
}

\author{Satpreet H Singh$^{1*}$}
\author{Kevin Jiang $^1$}
\author{Kanchan Bhasin$^2$}
\author{Ashutosh Sabharwal$^{2}$$^\dagger$}
\author{Nidal Moukaddam$^{1}$$^\dagger$}
\author{Ankit B Patel$^{1,2}$$^\dagger$}

\begin{abstract}
% [165 words (ideally 150)]
% \linenumbers
% \begin{linenumbers}
Semi-structured interviews (SSIs) are a commonly employed data-collection method in healthcare research, offering in-depth qualitative insights into subject experiences. 
Despite their value, the manual analysis of SSIs is notoriously time-consuming and labor-intensive, in part due to the difficulty of extracting and categorizing emotional responses, and challenges in scaling human evaluation for large populations. 
In this study, we develop RACER, a Large Language Model (LLM) based expert-guided automated pipeline that efficiently converts raw interview transcripts into insightful domain-relevant themes and sub-themes.
We used RACER to analyze SSIs conducted with 93 healthcare professionals and trainees to assess the broad personal and professional mental health impacts of the COVID-19 crisis.
RACER achieves moderately high agreement with two human evaluators (72\%), which approaches the human inter-rater agreement (77\%).
Interestingly, LLMs and humans struggle with similar content involving nuanced emotional, ambivalent/dialectical, and psychological statements. 
Our study highlights the opportunities and challenges in using LLMs to improve research efficiency and opens new avenues for scalable analysis of SSIs in healthcare research.
% \end{linenumbers}
\end{abstract}

\begin{keyword}
Semi-structured interviews \sep
Large language models \sep 
Healthcare NLP \sep
Burnout \sep
COVID-19 
\end{keyword}
\end{frontmatter}

\extrafootertext{ 
$^1$Baylor College of Medicine, Houston, TX; 
$^2$Rice University, Houston, TX; 
$^*$ Corresponding author (satpreetsingh@gmail.com);
$^\dagger$ Equal contribution
}

%%%%%%%%%%%%%%%%%%%%%%%%%%%%% #############################
\section{Introduction}

% BG on SSIs + motivation for RACER
Semi-structured interviews (SSIs) are a widely used qualitative research method in healthcare research, that provide an in-depth understanding of subjects' experiences in their own words \cite{adams2010joys}.
% more BG on SSIs
SSIs require interviewers to ask prespecified `root' questions, along with the option to ask follow-up questions to gain clarity on the interviewee's responses.
This flexibility is a key characteristic of SSIs, allowing for a more dynamic and responsive data collection process, especially in areas where exploratory forays are needed. 
The adaptability of SSIs is particularly beneficial in exploring complex or sensitive topics such as mental health.
SSIs allow rapport building between interviewer and subject, and facilitate candid responses on sensitive matters. 
The open-ended nature of follow-up questions gives subjects the freedom to reflect on experiences and articulate thoughts without judgement.
This helps reveal the nuances, contradictions, and diversity of perspectives that traditional fixed quantitative surveys may overlook. 
However, the traditional manual analysis of these interviews is a time-consuming and resource-intensive process. 
The advent of Large Language Models (LLMs), such as GPT-4 \cite{gpt4,lee2023benefits,lee2023harnessing}, offers a novel and efficient method to extract and interpret data from such text corpora. 
Yet, the validity of LLMs in analyzing emotional states may be limited in circumstances where participants express multiple emotions or conflicting (dialectical) states. 

% BG re SSI dataset + importance of working on SSI problem
As a case-study, we leveraged data from SSIs, conducted during the peak of the COVID-19 crisis in 2020, to understand the mental well-being of 93 healthcare professionals and trainees. 
The COVID-19 pandemic brought to the forefront profound personal and professional challenges experienced by healthcare workers. 
Fear of infecting family members, grief over patient deaths, moral dilemmas in resource allocation, and anxieties about professional preparedness collectively introduced a heightened level of psychological complexity and stress in the lives of healthcare professionals.
The stigma surrounding the pursuit of mental health support exacerbated these challenges, leaving healthcare workers hesitant to openly discuss their difficulties or seek assistance.

% a summary of what's coming up in rest of paper
In this paper, we developed \textbf{RACER}, an expert-guided automated pipeline that \textbf{R}etrieved responses to about 40 questions per SSI, \textbf{A}ggregated responses to each question across all subjects, \textbf{C}lustered these responses for each question into insightful domain-relevant \textbf{E}xpert-guided themes~\cite{lee_harnessing_2023}, and finally \textbf{R}e-clustered responses to produce a robust result. 
% Automated response validation and re-tries were added to each stage of the pipeline to avoid errors propagating downstream. 
Human evaluation on a subset of the total population revealed moderately high agreement \cite{mchugh2012interrater} between humans and RACER outputs, and similarities between inter-human disagreement and human-machine disagreement. 
% We also quantify and analyse sources of disagreement. 
We summarize our findings from applying RACER to our SSI-survey on the experiences of healthcare professionals and trainees during COVID-19, to reveal the power of this approach.
Our results demonstrate both the capabilities and the limitations leveraging LLMs to efficiently process and extract insights from a large corpus of SSIs. 

%%%%%%%%%%%%%%%%%%%%%%%%%%%%% #############################
\section{Results}
\label{sec_results}

\begin{table}[tbhp!]
    \centering
    \caption{Demographic Characteristics of the Study Population}
    \begin{tabular}{|l|c|}
    \hline
    \textbf{Characteristic} & \textbf{Percentage} \\
    \hline
    Gender & \\
    \quad Male & 54.84\% \\
    \quad Female & 45.16\% \\
    \hline
    Age Group & \\
    \quad 22-33 years & 39.78\% \\
    \quad 34-45 years & 32.26\% \\
    \quad 46-60 years & 16.13\% \\
    \quad 61+ years & 5.38\% \\
    \quad Unclear/Excluded & 6.45\% \\
    \hline
    Healthcare Professional/Student Type & \\
    \quad Physicians & 54.84\% \\
    \quad Medical Students & 21.51\% \\
    \quad Nurses & 8.60\% \\
    \quad Residents & 7.53\% \\
    \quad Other Professionals & 12.90\% \\
    \quad Unclear/Excluded & 1.08\% \\
    \hline
    Location & \\
    \quad Houston, Texas & 44.09\% \\
    \quad Other Texas & 21.50\% \\
    \quad Florida & 10.75\% \\
    \quad Mid-West US & 13.98\% \\
    \quad Other US & 5.38\% \\
    \quad Unclear/Excluded & 2.15\% \\
    \hline
    Marital Status & \\
    \quad Not married & 41.94\% \\
    \quad Married & 52.69\% \\
    \quad Unclear/Excluded & 5.38\% \\
    \hline
    Have Kids? & \\
    \quad Yes & 51.61\% \\
    \quad No & 45.16\% \\
    \quad Unclear/Excluded & 3.23\% \\
    \hline
    % Caretaking Responsibilities & \\
    % \quad Family Members & 15.05\% \\
    % \quad Animals & 1.08\% \\
    % \quad Partial Caretakers & 4.30\% \\
    % \quad Financial Support & 1.08\% \\
    % \quad None & 75.27\% \\
    % \quad Unclear/Excluded & 1.08\% \\
    % \hline
    % Year of Study (Students) & \\
    % \quad First Year & 20.00\% \\
    % \quad Second Year & 40.00\% \\
    % \quad Third Year & 25.00\% \\
    % \quad Fourth Year & 10.00\% \\
    % \quad Excluded & 5.00\% \\
    % \hline
    % Training Institution & \\
    % \quad Baylor College of Medicine & 24.73\% \\
    % \quad University of Texas & 13.98\% \\
    % \quad Texas Institutions & 24.73\% \\
    % \quad Multiple Institutions & 41.94\% \\
    % \quad Outside US & 18.28\% \\
    % \quad Unspecified/Missing & 3.23\% \\
    % \hline
    % Physician Training Location & \\
    % \quad In US & 90.74\% \\
    % \quad Not in US & 1.85\% \\
    % \quad Unclear & 7.41\% \\
    % \hline
    Specialty Area & \\
    \quad Emergency Medicine & 26.88\% \\
    \quad Psychiatry & 16.13\% \\
    \quad Pulmonary Critical Care & 16.13\% \\
    \quad Internal Medicine & 11.83\% \\
    \quad Neurology/Neurocritical Care & 5.38\% \\
    \quad Surgery/ER & 5.38\% \\
    \quad Pediatrics & 5.38\% \\
    \quad Other Specialties & 17.22\% \\
    % \quad Cardiology/Respiratory & 4.30\% \\
    % \quad OBGYN & 2.15\% \\
    % \quad Infectious Diseases & 1.08\% \\
    % \quad Anesthesiology/Critical Care & 2.15\% \\
    % \quad Pathology & 3.23\% \\
    % \quad Head/Neck Surgery & 1.08\% \\
    % \quad Fertility & 1.08\% \\
    % \quad Oncology & 2.15\% \\
    \quad Unclear/Excluded & 2.15\% \\
    \hline
    Years of Practice & \\
    \quad Under 15 Years & 71.23\% \\
    \quad 15-30 Years & 20.55\% \\
    \quad Over 30 Years & 5.48\% \\
    \quad Unclear/Excluded & 1.37\% \\
    \hline
    \end{tabular}
    \label{tab:demographics}
\end{table}

\begin{figure*}[tbhp!]
\centering
\includegraphics[width=0.65\linewidth]{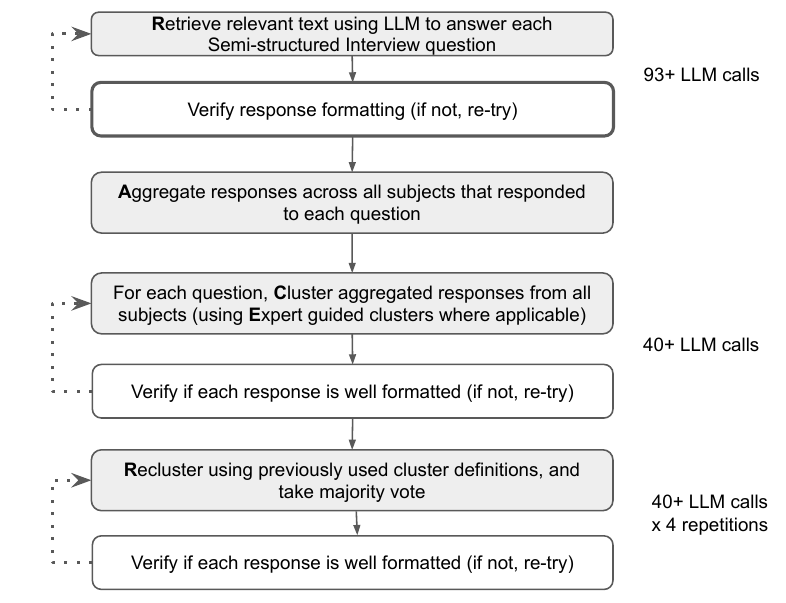}    
\caption{
\centering
\textbf{Stages of the RACER (Retrieve, Aggregate, Cluster with Expert guidance, and Re-cluster) Semi-Structured Interview (SSI) processing pipeline:}  
First, \textbf{R}etrieve relevant responses to each SSI question. 
\textbf{A}ggregate responses across subjects before \textbf{C}lustering them into themes (and subthemes) defined by \textbf{E}xperts. 
To assess robustness, \textbf{R}e-cluster multiple times and make assignments by majority vote. 
The pipeline efficiently converts SSI text into meaningful themes with confidence scores.
}
\label{fig1_pipeline}
\end{figure*}

%%%%%%%%%%%%%%%%%%%%%%%%%%%%% #############################
\subsection{Recruitment and interview of a diverse sample of healthcare professionals and trainees}
\label{subsec_recruit}

Healthcare professionals and trainees across different specialties and career stages were recruited via snowball sampling method~\cite{goodman1961snowball}, described as follows. 
The investigators asked colleagues if they knew of anyone willing to participate in interviews about their COVID-19 experiences. 
Announcements were also posted online and through professional networks. 
Participation was voluntary with no compensation provided. 
Approval was obtained from the Baylor College of Medicine (Houston, TX) Institutional Review Board.
The interviews were performed by a team of two research coordinators with healthcare backgrounds, and a third-year medical student, under the supervision of the investigators.

The study population of healthcare professionals and trainees consisted of 93 subjects (51 male, 42 female) with diverse demographics (Table 1). 
Subjects were from 22 years to over 61 years in age, and were located predominantly in Texas. 
Over half were married and had children. 
Most (75\%) had no care-taking responsibilities in addition to child-care. 
Professionally, the sample included physicians (54.8\%), medical students (21.5\%), nurses (8.6\%), residents (7.5\%) and other healthcare professionals. 
Subjects trained at multiple institutions, with prominent representation from Baylor College of Medicine and University of Texas systems. 
Various specialties were represented in the cohort, with emergency medicine, psychiatry and pulmonary/critical care among the most common.

SSIs were conducted over videoconferencing using a standard template consisting of a total of 41 questions, including four questions that were only asked to students, and seven questions that were asked to only non-students.
Questions were either \textit{factual}, concerning demographics and personal and professional background, or \textit{open ended}, where interviewees were asked to talk about their experiences during the COVID-19 pandemic, focusing on their exposure to the virus, work impacts, emotional responses, future outlooks, and coping strategies. 
Interviewees discussed how they had practiced in high-risk areas, their concerns for personal and family safety, and modifications made to their routines. 
They also reflected on the physical toll the crisis had taken. 
The impact on their work included changes in working hours, shifts in patient care quality, and altered management approaches. 
Emotional and psychological questions revealed how the crisis affected them emotionally, the level of support they received, family dynamics, and changes in burnout levels. 
Looking ahead, they pondered the crisis's short-term and long-term impacts on their careers and specialty choices. 
Finally, they shared their openness to seeking help for burnout or mental overwhelm and identified potential obstacles in obtaining this help.
Students were not asked clinical-practice related questions, and were instead asked about how their training was being affected by pandemic-related changes.
Interviews lasted on average 26.7 +/- 8.9 s.d. minutes.
When transcribed from raw interview audio into text transcripts (using Otter.AI\cite{otter}), were on average 4044.30 +/- 1348.34 s.d. words long.

%%%%%%%%%%%%%%%%%%%%%%
\begin{figure*}[tbhp!]
\centering
\includegraphics[width=0.80\linewidth]{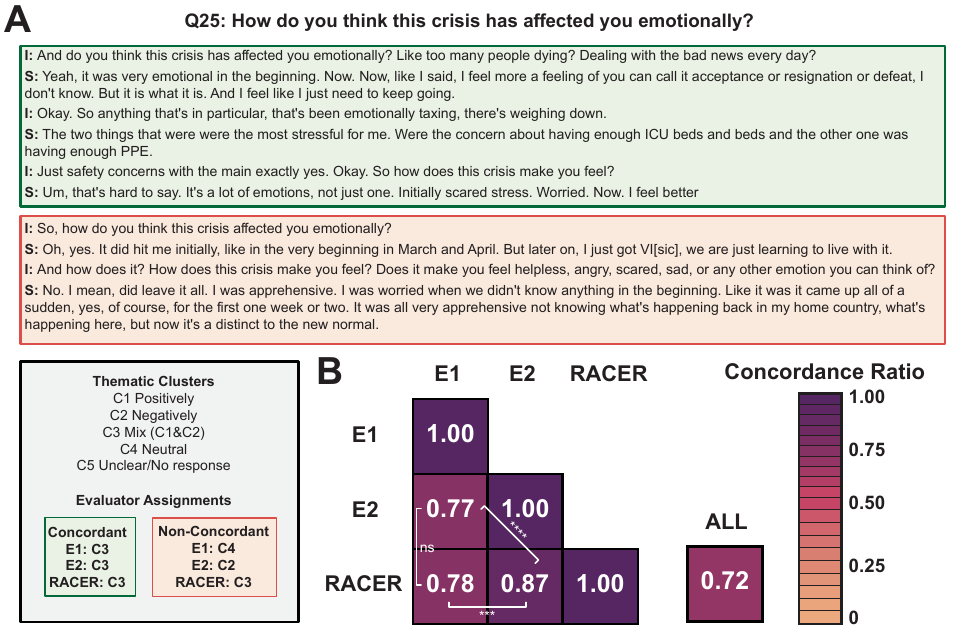}
% \includesvg[width=0.80\linewidth]{fig2_human_val_racer.svg}
\caption{
\textbf{Human-RACER approaches resembles human-human disagreement:} 
(A) Transcript segments from two different subjects being asked “How the [COVID-19] crisis has affected you emotionally” that were either all concordant or all non-concordant between evaluators, displaying the ambiguity that exists in parsing free response. 
(B) The concordance ratio calculated between evaluator pairs and between all evaluators. A chi-squared test with yates continuity correction between the three different evaluator pairings showed human evaluator concordance did not differ from evaluator one’s concordance with RACER. * p $<$ 0.5, ** p $<$ 0.01, *** p $<$ 0.001, **** p $<$ 0.0001. 
}
\label{fig2_humanval}
\end{figure*}

%%%%%%%%%%%%%%%%%%%%%%%%%%%%% #############################
\subsection{RACER extracts relevant interviewee responses and robustly clusters them}
\label{sec_pipeline}

We developed an LLM-based automated pipeline called \textbf{RACER} (Figure \ref{fig1_pipeline}) that converts a corpus of text SSI transcripts into insightful themes per interview question. 
RACER, consists of four stages, \textbf{R}etrieve, \textbf{A}ggregate, \textbf{C}luster with \textbf{E}xpert guidance, and \textbf{R}ecluster:

\emph{Retrieve}: 
We first structured interview transcripts by using an LLM (OpenAI's GPT-4\cite{gpt4}) to \textit{retrieve}
relevant SSI text in response to each of the questions in the interview template. 
(See Methods LLM prompt details)
To avoid LLM `hallucinations' \cite{tonmoy2024comprehensive}, we asked the LLM to provide `evidence' in the form of text quoted verbatim from the transcript, to back up its response to each question.
LLM outputs missing either answers or backing evidence to any question were automatically detected and rerun.

\emph{Aggregate}: 
For each question, we then aggregated the retrieved responses across all subjects who were asked that question. 

\emph{Cluster with Expert guidance}: 
We then asked the LLM to \textit{semantically cluster} the  responses into primary and secondary clusters (`themes' and `sub-themes'). 
For most questions, we provided the LLM expert-guidance in the form of primary-cluster definitions. 
The LLM discovered secondary clusters (or sub-themes) automatically.
Expert-provided cluster definitions were always mutually exclusive and collectively exhaustive, while those discovered by the LLM were not constrained to be so.
Similar to before, invalid LLM responses, e.g. those missing cluster assignments for any subjects, were automatically re-run.

\emph{Re-Cluster}: 
Leveraging the probabilistic nature of LLMs, we assessed the \textit{robustness} of the clustering process by re-running it four more times, employing the same cluster definitions and validation criteria as in the initial step.
We used a majority vote over 5 runs to assign subjects to clusters, to get robust cluster assignments for all downstream processing.
The number of votes (3, 4 or 5 out of total 5 LLM calls) additionally provided a synthetic measure of LLM \textit{confidence} \cite{kompa2021second,tanneru2023quantifying}.
% sun2022quantifying,xiao2022uncertainty,jousselme2023uncertain,
Only a very small fraction of subject-question pairs (12 out of 3342, 0.36\%) had no `robust' cluster assignments after applying the majority voting process.

All together, we found that RACER was able to take unstructured transcriptions and extract relevant and insightful, clustered responses in a robust manner for downstream human analysis.

%%%%%%%%%%%%%%%%%%%%%%%%%%%%% #############################
\subsection{
Human-machine disagreement approaches inter-human disagreement}
\label{sec_humanval}

To validate the output of running RACER on our SSI dataset, two human evaluators cross-checked the resulting cluster assignments for 20 randomly-selected subjects across 28 open ended questions. 
Using the same cluster definitions as were previously used by RACER, each human evaluator (E1 and E2) independently read the raw transcript file and assigned each subject's answers to the primary clusters.
Evaluator cluster assignments were then compared to RACER's robust cluster assignments. 
To quantify agreement, we defined a \textit{concordance score} and a \textit{concordance ratio} as follows:
If the clusters for a given subject-question pair matched exactly (for mutually exclusive clusters), or were sub-sets or super-sets of each other (for mutually non-exclusive clusters) they were assigned a concordance score of 1.
Conversely, mismatch was assigned a concordance score of 0. 
The overall concordance ratio is the proportion of matched subject-question pairs between evaluators. 

% Across all reviewed subjects, there was high concordance (77\% with E1, 87\% with E2) between each human evaluator and RACER, and equally high concordance between the two human evaluators (Figure \ref{fig2_humanval}B).
We observed a concordance of 78\% (E1) and 87\% (E2) between human evaluators and RACER, and a 77\% (E1-E2) inter-rater concordance (Figure \ref{fig2_humanval}B).
When all three evaluators were compared simultaneously, there was a little decrease in the concordance (72\%), indicating that across the majority of subject-question pairings, cluster assignments produced by humans and RACER were all in agreement. 
% However, none of the compared pairs (E1-E2, E1-LLM, E2-LLM) were perfectly consistent with each other, highlighting the ambiguity of parsing free responses from human subjects as exemplified in Figure \ref{fig2_humanval}A. 
See Methods for additional details.

\subsection{Machine "confusion" resembles human confusion}
\label{subsec_confidence}

\begin{figure*}[tbhp!]
\centering
\includegraphics[width=0.95\linewidth]{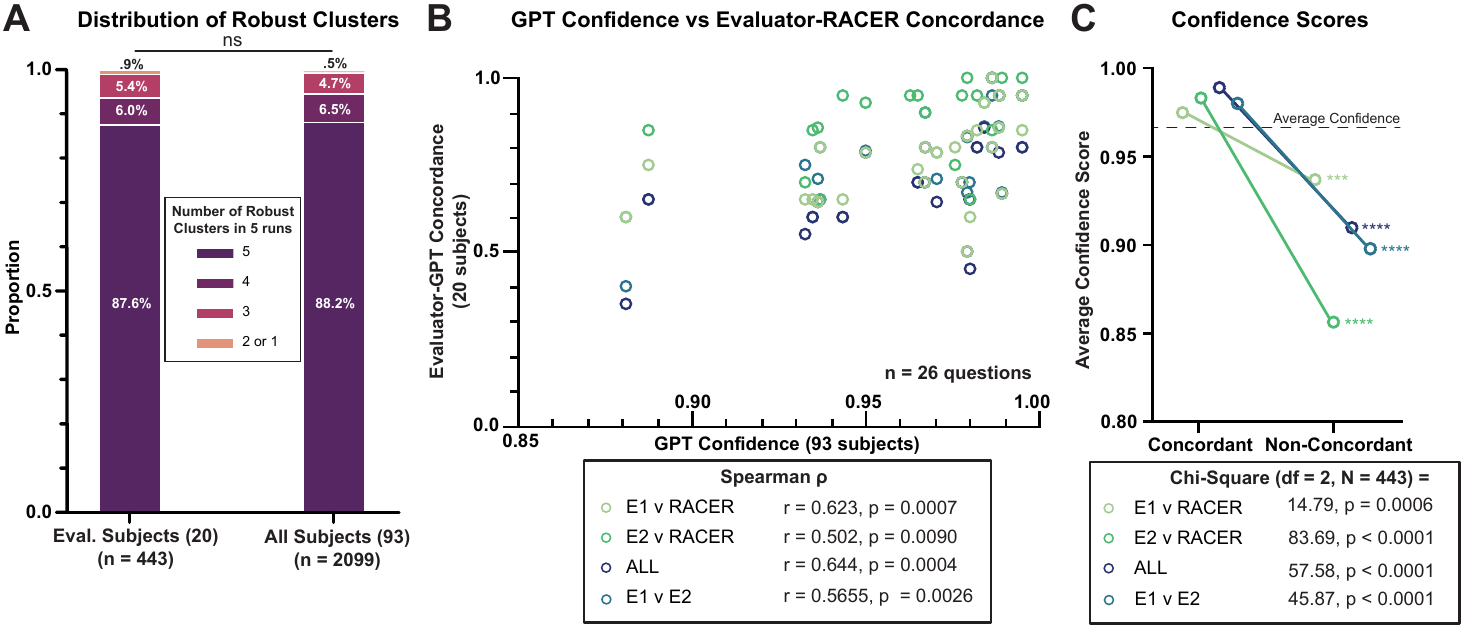}
% \includesvg[width=0.95\linewidth]{fig3_confidence_racer.svg}
\caption{
%\centering{
%Analysis of RACER confidence scores: 
%RACER repeats the clustering LLM call 5 times to estimate the robustness (or confidence) of the cluster assignments. 
%(A) we see that the overwhelming majority of the assignments are robust across calls,
%(B) Evaluator-GPT concordance and RACER confidence are highly correlated, and
%(C) RACER showed statistically significant drops in confidence between concordant and non-concordant subject-question pairs.  
\textbf{RACER “confidence” correlates with evaluator concordance and reveals areas of human disagreement:} 
(A) Distribution of the proportion of subject-question pairs that were clustered robustly across subject-question pairs human evaluators examined (20 subjects) or across all subject question pairs (93 subjects). 
(B) Average RACER confidence scores for all subjects (n = 93) for a given question correlate significantly with the evaluator pair concordance (n = 20) using Spearman Rank. (C) Average RACER confidence scores calculated within concordant vs non-concordant subject-question pairs between evaluators. Chi-square test was conducted to determine if distribution of confidence scores differed between concordant vs non-concordant subject-question pairs.Correlation. * p $<$ 0.5, ** p $<$ 0.01, *** p $<$ 0.001, **** p $<$ 0.0001. 
%}
}
\label{fig_confidence}
\end{figure*}

We examined the confidence score produced by RACER per subject-question pair to see how it might affect the subject-question pair's concordance with human evaluators (Figure \ref{fig_confidence}).
% In the process of generating robust clusters with the LLM, we examined how the degree of robustness (assigning the same cluster to a given subject-question pair 5 times) may affect it's concordance with human evaluators. 
Amongst the 443 subject-question pair sample evaluated by humans, 392 (87.7\%) were entirely robust (5 of 5 repeat clusters).
This was similar to the population confidence score distribution 88.2\% (1852 of 2099 subject-question pairs). 
RACER's average confidence across all subjects for a given question showed significant and positive correlations with Human-RACER concordance of the 20 subjects evaluated for those same questions.
% , which was representative of the entire confidence across all subject-question pairs (1852 of 2099, 88.2\%; 
Additionally, we observed that the confidence scores of subject-question pairs that were concordant between human evaluators and RACER, were higher than for non-concordant subject-question pairs. This was due to significant differences in the proportion of lower confidence subject-question pairs between concordant and non-concordant groups. 
% However, when we accounted for the difference in concordant vs non-concordant sample size, an interesting trend emerged. 
% While between individual evaluators and the LLM there was a mixed change in scaled confidence, when both human evaluators were compared, the difference between coherent and non-coherent scaled confidence was much greater (Appendix \ref{Sup3_confidence}). 
% This indicated that there were non-overlapping subject-question pairs of lower robust clustering where human evaluators disagreed with the LLM, suggesting areas of ambiguity. 
% Interestingly, when we applied the LLM confidence to human-human concordance, similar trends emerged. 
Interestingly, when we juxtaposed RACER confidence scores against human-human concordance, we observed that RACER confidence was lower when humans were non-concordant. 
This suggests that areas where RACER was less confident or `confused' were also areas where human evaluators tended to disagree. 
Thus the RACER confidence generated via repeated clustering could also serve as an indicator of SSI ambiguity or difficulty.
% This suggested that subject-question pairs where the LLM was less robust in its clustering were areas where human evaluators were also more likely to disagree, indicating areas of confusion for human evaluators were areas the LLM faltered as well. 
% Additional analysis of how the LLM confidence and concordance might correlate showed that the average confidence the LLM had across all subjects for a given question showed significant and positive correlations with the human-LLM concordance of the 20 subjects evaluated of those same questions (Appendix \ref{Sup3_confidence}).

%%%%%%%%%%%%%%%%%%%%%%%%%%%%% #############################
\subsection{Insights using RACER on healthcare worker experience during COVID-19}

We now summarize the insights gleaned from analyzing SSIs with 93 subjects using our automated processing pipeline. 

\subsubsection{COVID-19 exposure, response, work impact and work changes}
\label{sec_response}

% **COVID-19 Exposure and Response**
% - Q12: Over the past two months, have you practiced clinically in areas where you could be in touch with patients who have covid-19? [ONLY nonstudents]
% - Q13: Are you concerned about your safety, and how?
% - Q14: Are you concerned about the safety of loved ones, and how?
% - Q15: Have you modified your routine to protect yourself or others, and how?
% - Q16: Has this crisis taken a toll on you physically in any way?
% - Q23: How prepared do you feel for the COVID-19 pandemic on a personal level?
% - Q24: How prepared do you feel the unit/hospital is for the COVID-19 pandemic?

% **Work Impact and Changes**
% - Q17: How many hours are you working on average (per week) nowadays?
% - Q18: How has your working schedule and logistics changed?
% - Q19: How do your working hours compare to pre-covid-19 crisis?
% - Q20: How do think the covid-19 crisis has affected the quality of patient care?
% - Q21: How has it changed your approach to management? (different from usual, at odds with existing guidelines, may not be as effective, etc.) [ONLY nonstudents]
% - Q22: Are your processes different for end-of-life decisions? Do you have to take people off ventilator more frequently? [ONLY nonstudents]
% - Q30: How do you feel about working from home OR at the frontlines? [ONLY nonstudents]
% - Q31: Do you feel you should be able to handle this as a healthcare professional? [ONLY nonstudents]

\begin{figure*}[htbp]
    \centering
    % First row
    \begin{minipage}[b]{0.48\linewidth}
        \labeledtikzfig{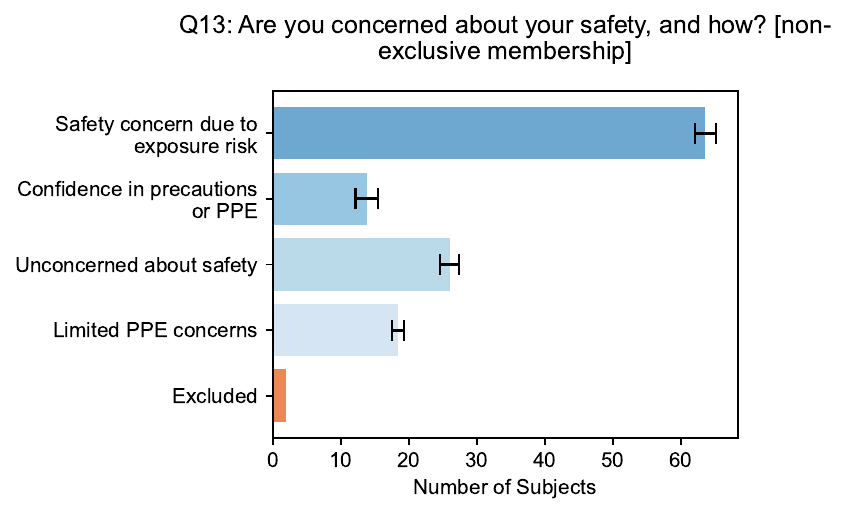}{A}{-0.05}{1.1}
    \end{minipage}
    \hfill
    \begin{minipage}[b]{0.48\linewidth}
        \labeledtikzfig{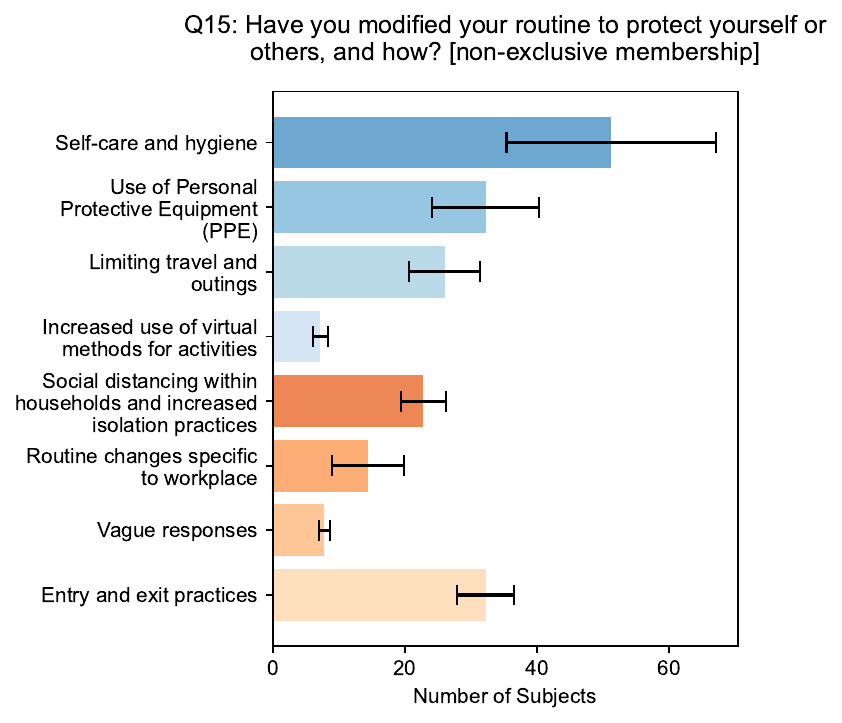}{B}{-0.05}{1.1}
    \end{minipage}
    % Second row
    \begin{minipage}[b]{0.48\linewidth}
        \labeledtikzfig{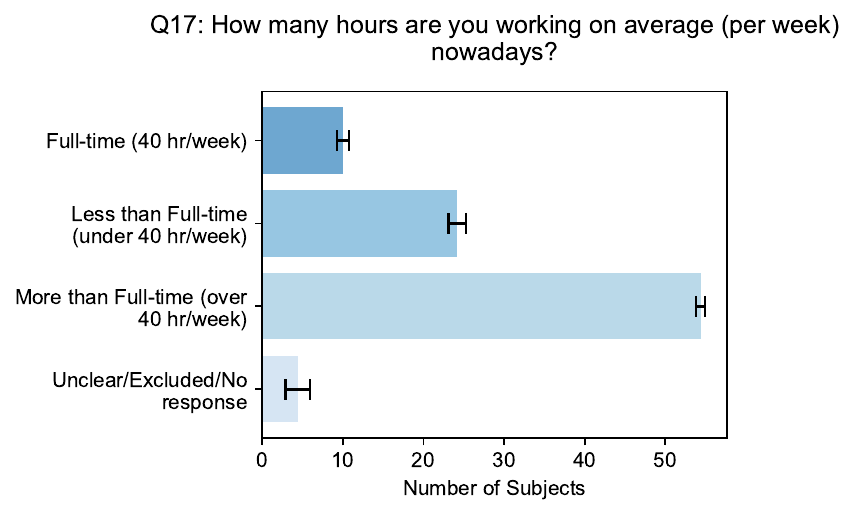}{C}{-0.05}{1.1}
    \end{minipage}
    \hfill
    \begin{minipage}[b]{0.48\linewidth}
        \labeledtikzfig{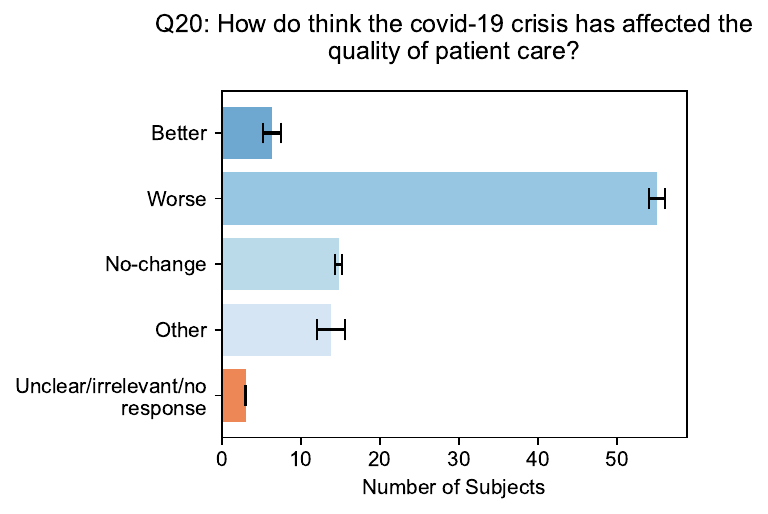}{D}{-0.05}{1.1}
    \end{minipage}
    % Third row
    \begin{minipage}[b]{0.48\linewidth}
        \labeledtikzfig{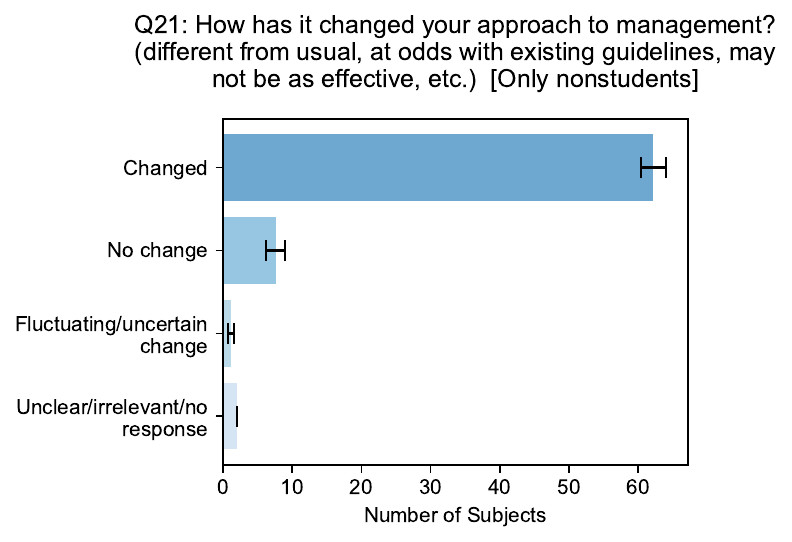}{E}{-0.05}{1.1}
    \end{minipage}
    \hfill
    \begin{minipage}[b]{0.48\linewidth}
        \labeledtikzfig{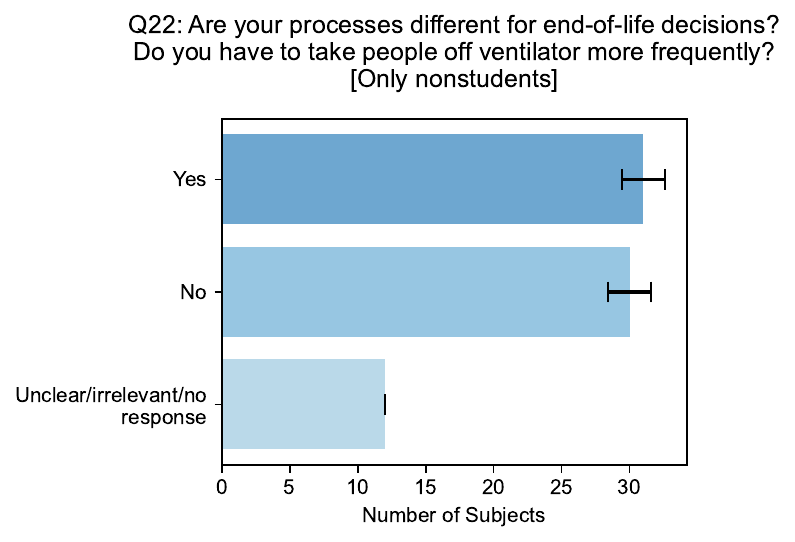}{F}{-0.05}{1.1}
    \end{minipage}

    \caption{Aggregated interview responses to selected questions about safety concerns arising from COVID-19 exposure, work impact, and medical management decisions. Error bars reflect cluster-assignment variability arising from re-clustering step in RACER.}
    \label{fig_work_impact}
\end{figure*}

The vast majority of practicing healthcare professionals reported having professional contact with COVID-19 patients in the past two months. Most subjects expressed safety concerns for themselves and loved ones, especially regarding viral exposure risks. Common protective measures adopted included heightened hygiene practices, using personal protective equipment, limiting travel and social interactions, and modifying routines at work and home to minimize transmission risks. Over half of the subjects reported physical tolls from the crisis, frequently citing exhaustion, disturbed sleep, and dietary changes (Figure \ref{fig_work_impact}). 

Most subjects felt personally prepared to handle the pandemic, attributing this largely to their medical knowledge, experience, and ability to adapt. Assessment of institutional preparedness was more varied, with around 60\% expressing their hospital/unit was prepared, but around 25\% felt improvements were still needed.

Working hours markedly increased for most subjects during the pandemic, with over 80\% reporting working more than 40 hours per week compared to pre-COVID times. For many, this resulted from escalations in patient load and administrative duties. Approaches to patient management also evolved, with the vast majority of practicing healthcare professionals stating their methods differed from usual practices. This included increased reliance on technology, more precautions with patients, and adjustments to treatments due to COVID-19. Most still felt capable of handling the situation professionally, though some desired more protections and support systems.

Among students and trainees, the majority believed they adhered closely to the Hippocratic oath during the pandemic. Their views on their educational institution's policies regarding medical students' roles during that time were divided, with half in agreement and others expressing mixed or negative sentiments, reflecting a spectrum of perspectives on the adequacy and effectiveness of institutional responses to the crisis.

%%%%%%%%%%%%%%%%%%%%%%%%%%%%% #############################
\subsubsection{Emotional and psychological impact, and support and coping strategies}
\label{sec_emo}

% **Emotional and Psychological Impact**
% - Q25: How do you think this crisis has affected you emotionally?
% - Q26: Do you feel supported by peers and/or family during this time?
% - Q27: Have you had more problems with family during this time?
% - Q28: Before this crisis, if someone asked you about your burnout level, what would you have answered?
% - Q29: How burned out do you feel nowadays (during the ongoing COVID crisis)?
% **Support and Coping Strategies**
% - Q34: Would you seek help if you felt burned out? How?
% - Q37: Would you get (professional or other) help/care if you felt mentally overwhelmed? How? When?
% - Q38: Any obstacles you foresee in getting help if you needed to?

\begin{figure*}[tbhp!]
    \centering
    % First row
    \begin{minipage}[b]{0.45\linewidth}
        \labeledtikzfig{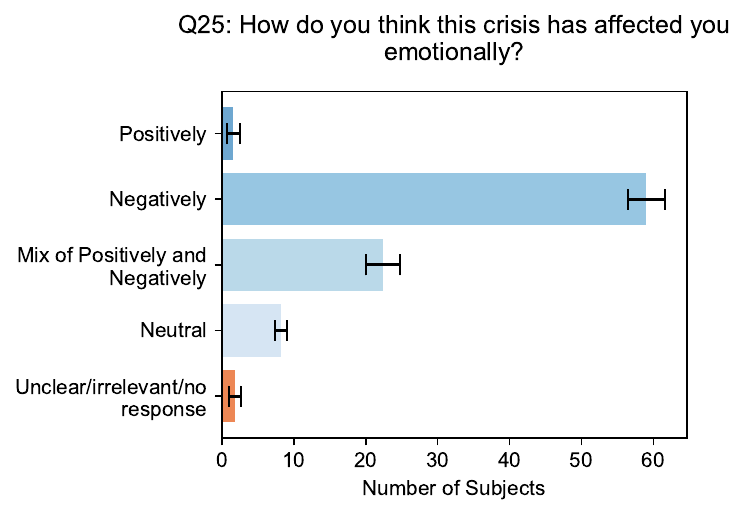}{A}{-0.05}{1.1}
    \end{minipage}
    \hfill
    \begin{minipage}[b]{0.45\linewidth}
        \labeledtikzfig{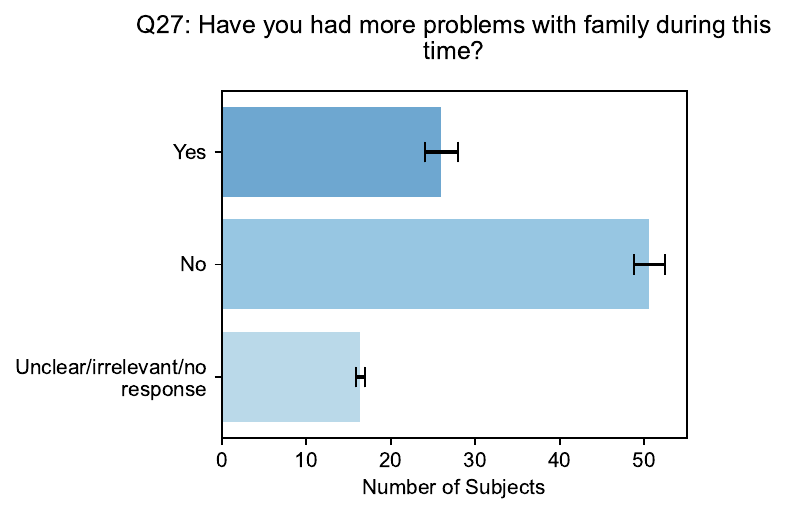}{B}{-0.05}{1.1}
    \end{minipage}
    
    % Second row
    \begin{minipage}[b]{0.45\linewidth}
        \labeledtikzfig{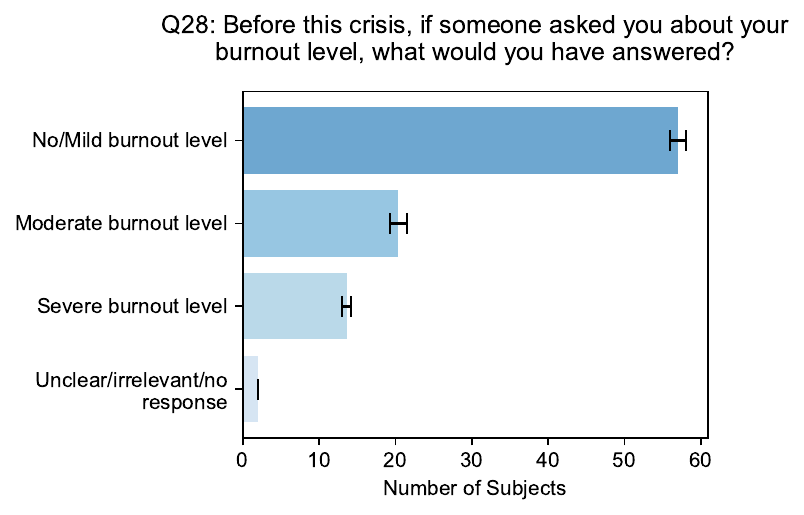}{C}{-0.05}{1.1}
    \end{minipage}
    \hfill
    \begin{minipage}[b]{0.45\linewidth}
        \labeledtikzfig{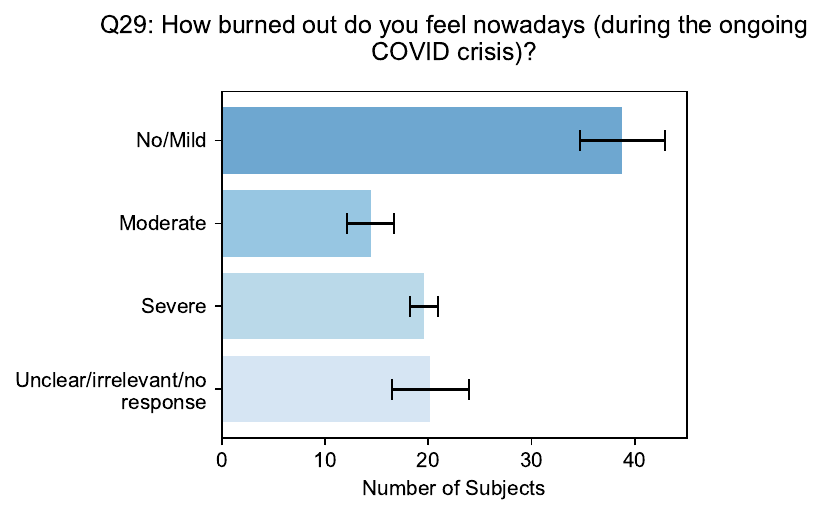}{D}{-0.05}{1.1}
    \end{minipage}
    
    % Third row
    \begin{minipage}[b]{0.45\linewidth}
        \labeledtikzfig{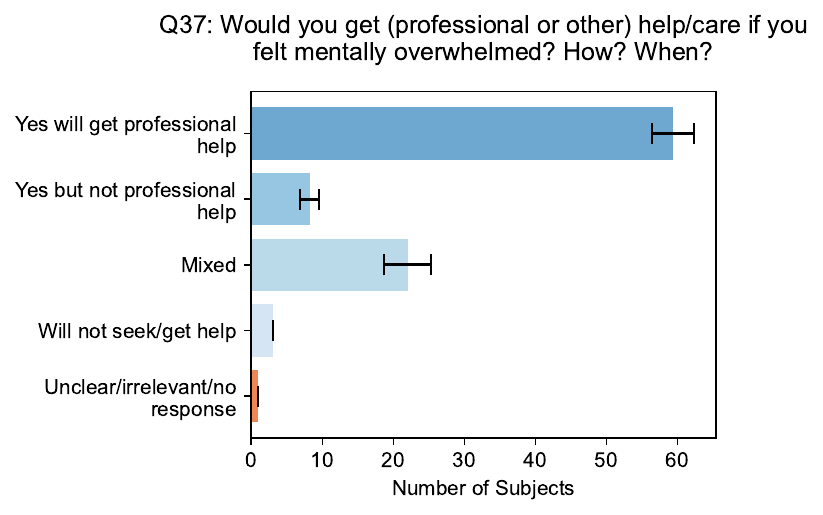}{E}{-0.05}{1.1}
    \end{minipage}
    \hfill
    \begin{minipage}[b]{0.45\linewidth}
        \labeledtikzfig{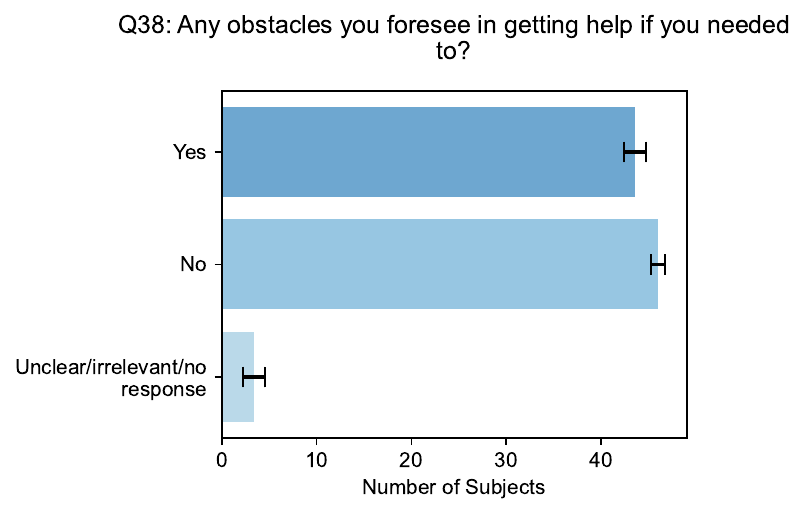}{F}{-0.05}{1.1}
    \end{minipage}

    \caption{
    Aggregated interview responses to selected questions about emotional and psychological impact, and support and coping strategies. 
    Error bars reflect cluster-assignment variability arising from re-clustering step in RACER.
    }
    \label{fig_emo}
\end{figure*}

The COVID-19 crisis negatively affected the emotional state of most subjects, with many reporting feelings of anxiety, stress, sadness, or anger. However, around 25\% indicated a mix of both positive emotions like gratitude as well as negative feelings. 
Despite those challenges, the overwhelming majority felt supported by peers and family, suggesting strong social networks within and outside the workplace. 
Family dynamics had been affected for some, with around a quarter reporting increased family problems during the pandemic. 
This data underscored the profound emotional and psychological effects of the crisis on healthcare professionals, juxtaposed with the resilience and support systems that helped them navigate these challenges. 

In regards to burnout, over 60\% of subjects assessed their pre-pandemic burnout as low or mild. When asked about current burnout, around 40\% still reported mild or no burnout, but the percentage reporting severe burnout rose from around 15\% pre-pandemic to 20\% during the crisis.
If feeling burned out, nearly 90\% stated they would seek help, with most mentioning professional resources like counseling. Over 60\% also reported they would seek professional help if feeling mentally overwhelmed, with therapists and workplace programs being commonly cited options. However, around 45\% still anticipated obstacles in getting help, including logistical barriers and stigma concerns (Figure \ref{fig_emo}).

%%%%%%%%%%%%%%%%%%%%%%%%%%%%% #############################
\subsubsection{Future considerations and professional outlook}
\label{sec_prof}

% **Future Considerations and Professional Outlook**
% - Q32: What impact do you see this crisis having on you in the near future?
% - Q33: What impact do you see this crisis having on you about five years from now?
% - Q35: Would you change jobs or career trajectories? [ONLY nonstudents]
% - Q36: Has this crisis affected your specialty decision or career plans in any way? [ONLY students]
% - Q39: If student or trainee, how closely do you feel that you are adhering to the Hippocratic oath during this time? [ONLY students]
% - Q40: If student or trainee, do you agree with your school's policies regarding medical students' roles at this time? [ONLY students]

\begin{figure*}[tbhp!]
    \centering
    % First row
    \begin{minipage}[b]{0.48\linewidth}
        \labeledtikzfig{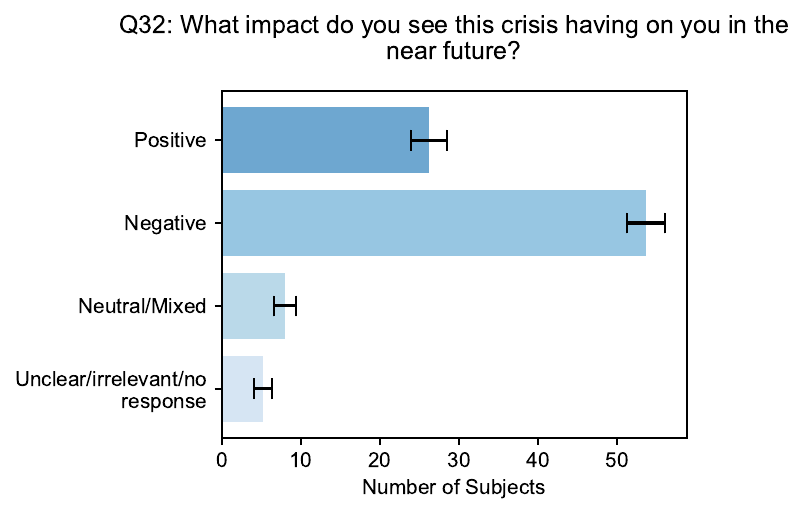}{A}{-0.05}{1.1}
    \end{minipage}
    \hfill
    \begin{minipage}[b]{0.48\linewidth}
        \labeledtikzfig{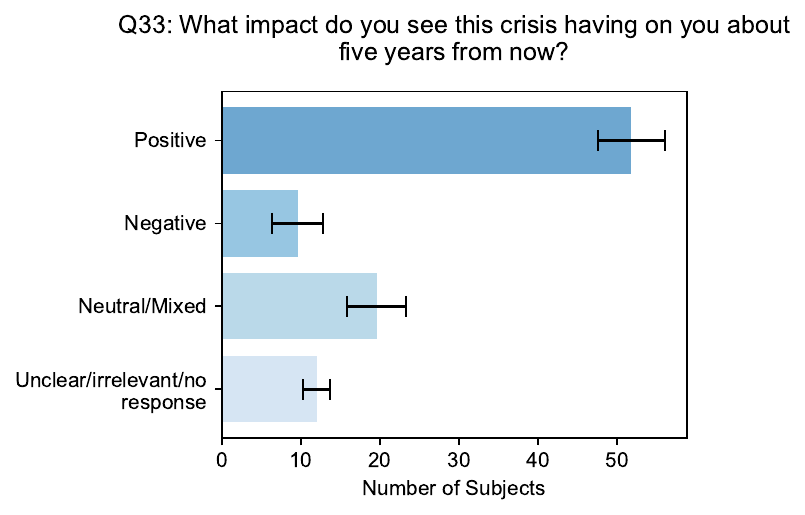}{B}{-0.05}{1.1}
    \end{minipage}
    
    % Second row
    \begin{minipage}[b]{0.48\linewidth}
        \labeledtikzfig{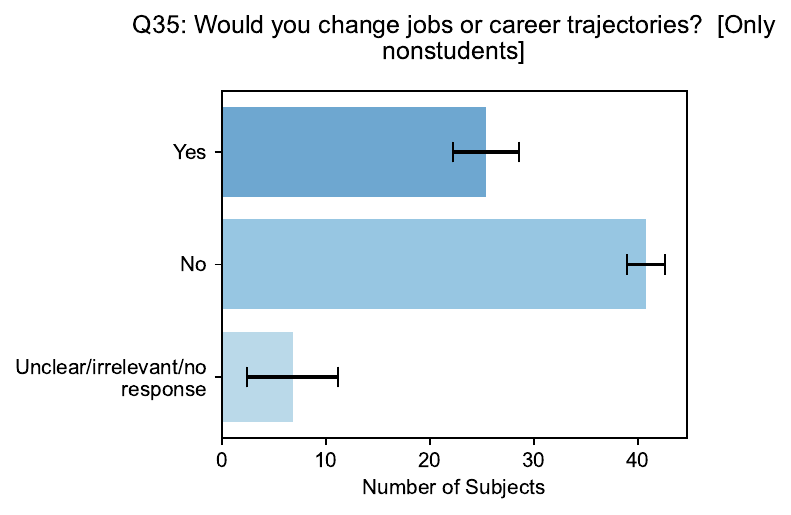}{C}{-0.05}{1.1}
    \end{minipage}
    \hfill
    \begin{minipage}[b]{0.48\linewidth}
        \labeledtikzfig{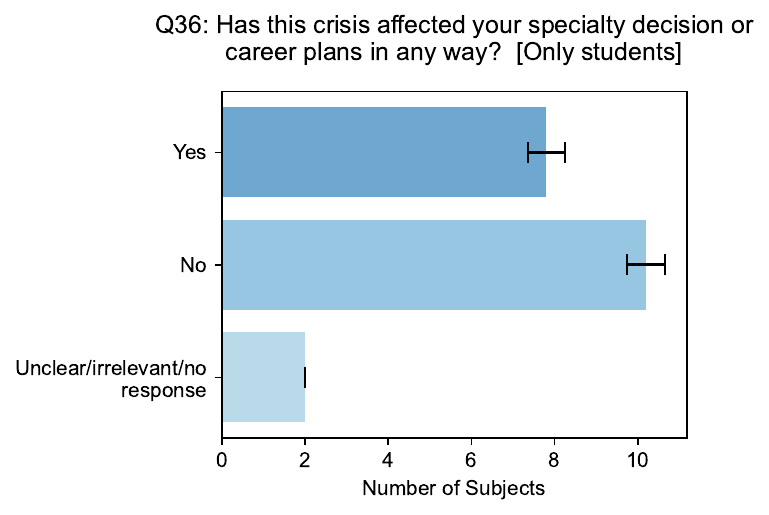}{D}{-0.05}{1.1}
    \end{minipage}

    \caption{Aggregated interview responses to selected questions about future considerations and professional outlook, as it relates to working in healthcare during or after the pandemic. Error bars reflect cluster-assignment variability arising from re-clustering step in RACER.}
    \label{fig_future}
\end{figure*}

When asked about near-term impacts, over 50\% expressed concerns about anticipated difficulties, health risks, economic instability, and significant lifestyle changes. However, around 15\% hoped for new opportunities and growth resulting from the crisis. Looking 5 years ahead, around 20\% expected advancements in healthcare practices and systems due to learned lessons. Though nearly 10\% feared lingering personal and professional impact.
Among non-students considering job changes, around 15\% expressed an immediate willingness to switch fields while around 18\% would change contingent on worsening conditions. 

Regarding effects on career plans, 35\% of students reported the crisis has impacted their specialty choices or work preferences. Specifically, around 20\% described reconsidering their specialty choice due to the pandemic. Another 15\% mentioned shifting their preferences regarding research involvement, practice locations, and other factors. However, 50\% of students stated the crisis has not affected their professional plans or specialty decisions.
Over 50\% of students explicitly stated adherence to their Hippocratic oath obligations, while 10\% conveyed adherence through descriptions of their clinical actions and interventions.
Of students agreeing with their school's pandemic policies, 40\% expressed unqualified agreement and 10\% provided positive justifications. However, around 15\% agreed tentatively due to concerns over student safety and curriculum changes (Figure \ref{fig_future}). 

%%%%%%%%%%%%%%%%%%%%%%%%%%%%% 
\section{Discussion}

\subsection{Summary}
Our study demonstrates the utility of RACER for efficiently analyzing semi-structured interviews (SSIs), particularly those exploring complex mental health topics within the healthcare domain. 
We introduce a novel approach by employing RACER to analyze emotions and psychological behaviors, opening new possibilities for exploration in mental health.
By providing expert-guided constraints and using automated response validation steps, RACER accurately extracts and robustly clusters relevant responses from interview transcripts. 
Automating these laborious manual tasks significantly enhances the scalability of SSI analysis.
The inter-rater agreement between LLM-assigned clusters and human expert clusters further bolsters our claims. 
The automated pipeline achieved moderately high concordance compared with manual evaluation by human annotators. 
The overall concordance ratio of 0.72 for RACER versus both human evaluators approaches the 0.77 concordance ratio between the two human evaluators.

\subsection{Limitations and tradeoffs}
Our findings reveal both the promises and current pitfalls of LLMs for SSI analysis.
We found that when the RACER struggled with robust clustering, both humans and machine were more likely to be non-concordant which could suggest shared limitations in handling complex emotions or psychologically nuanced statements \cite{boag_hard_2021} or ambiguity of the underlying SSI.
This underscores the indispensable role of human expertise in reviewing and interpreting LLM outputs, where RACER's confidence levels can guide expert scrutiny.

While RACER provided evidence in the form of quoting relevant interview text to support its response in the Retrieval step, the underlying methodology remains opaque. 
In contrast, human evaluators were able to describe their techniques, even if subjective. 
For instance, humans considered different amounts of contextual information outside the question scope, and inferred subject intentions to varying degrees, i.e. whether the subject needed to explicitly say certain phrases, or if they could be inferred from previous statements or knowledge of the subject matter.
An LLM's ability to consider large amount of contextual information can be a double-edged sword; beneficial if relevant information appears elsewhere in the transcript, but misleading if the research is indeed directed towards a narrow window of text around the question.

We demonstrated that LLMs can help discover knowledge by automatically extracting themes and topics from subject responses. 
However, good performance requires clear, mutually exclusive category definitions. 
We found it highly useful to involve domain experts early to precisely define mutually exclusive thematic clusters. 
For certain questions, where succinct mutually exclusive categorization was not possible, we chose to use LLM-discovered clusters.
However, validation of such non-exclusive categorization is challenging. 
Our results showed higher LLM accuracy and inter-rater agreement for questions with non-overlapping expert-defined clusters versus those allowing multiple clusters.

Additionally, human evaluators exhibited biases, such as default cluster tendencies requiring countering evidence (e.g. starting from a default of `no' and requiring evidence to switch to a `yes', or vice versa). 
Thus, expert human analysis also demonstrates cognitive variability and individual biases. 
Rather than definitive classifications, both human and machine outputs should be considered informed yet inherently biased perspectives on complex qualitative responses \cite{atari_which_2024}.
Thus, in the future, clearly delineating the parameters of evaluations with humans and RACER may improve concordance. 

While RACER's cluster assignments may deviate slightly from human reviewers, RACER was internally consistent and demonstrated high clustering repeatability for most questions.
Furthermore, unlike humans, RACER was able to efficiently process an extensive dataset of 93 subjects and can scale to significantly larger data set sizes that would otherwise be infeasible for human evaluators to handle.

\subsection{Future work}
For researchers undertaking projects in this emerging domain, both optimism and caution are warranted \cite{badal_guiding_2023,dash_evaluation_nodate,chiu_computational_2024,tang_evaluating_2023, wornow_shaky_2023, shah_creation_2023}.
With appropriate constraints and validation, LLMs can accelerate knowledge extraction from SSIs.
We implemented safeguards against hallucination risks like requiring verbatim textual evidence for an answer, which constrained the LLM to mostly avoid fabricating content.
While this is already an area of active research, the possibility of a few false positives remains and needs to be accounted for in downstream use. 

While evaluation of LLM outputs through comparison to multiple human raters was helpful through comparison to multiple human raters, inter-rater agreement must also be looked at to assess inherent ambiguity.
To further improve performance, we recommend specialized training for both SSI interviewers and human evaluators. 

We found it useful to generate an ensemble of LLM clustering outputs from repeated runs, and used it to extract robust cluster assignments and to get a measure of model uncertainty.
Future work exploring this direction could produce useful methods that help build trust in LLM-assisted analyses and inform human-in-the-loop processes for high-stakes applications \cite{bienefeld_solving_2023}.

%%%%%%%%%%%%%%%%%%%%%%%%%%%%% #############################
% \newpage

\section{Methods}

\subsection{Semi-structured interviews}
Interviewers were provided with a standard template to guide their discussions. 
The subjects were all healthcare professionals or trainees, including physicians, nurses, and medical students.
The interviews followed a semi-structured format, where the interviewers were instructed to cover a previously decided list of questions, and were allowed to ask exploration questions if the `root' question was not answered. The questions covered in the SSIs are listed in \ref{supp_prompt_retrieve}.
Raw audio / video interview files were transcribed into text format using the Otter.AI transcription service \cite{otter}. 
Out of 100 interviews conducted, 7 were compromised due to data-corruption/loss issues, providing a total of 93 transcriptions for further processing.
Voice to text transcription was carried out using Otter.AI\cite{otter}, which attempts to perform automated speaker diarization, but does not do so perfectly. 
To the best of our knowledge, this shortcoming did not seem to influence the subsequent processing steps.

\subsection{RACER}
We used the OpenAI GPT-4 LLM for all our work, except for prompts which exceeded GPT-4's limits, where we used GPT-4-32k.

\subsubsection{Retrieval}
In this step, the model was tasked with retrieving relevant responses for each question from a predefined list of questions (listed in \ref{supp_prompt_retrieve}) from the transcript. 
The prompt for the LLM consisted of instructions and a template consisting of the aforementioned list of questions and what format each question's response should be in, followed by the entire SSI transcript. 
The full prompt is detailed in \ref{supp_prompt_retrieve}.  

LLM Response Validation: 
By asking the LLM to respond in a structured format, we could partially automate the process of verifying the LLM's response.  
The LLM is called once for each subject, and then the response is parsed using the Python Pandas library. 
The LLM's response is marked invalid if it is ill-formatted (not parsable in tab-separated-values format) or incomplete (wrong number of rows, i.e. questions, or columns, i.e. incomplete response). 
The LLM is called again on invalid responses till the LLM returns a valid  response. 
We found that at most 4-5 (5\%-6\%) subjects would have invalid responses in the first attempt, and in total, we were making about 10\% additional calls to get valid responses for all subjects. 
The most common issues were that the LLM would sometimes be incomplete (skip questions, end output before final question) and sometimes use the specified tab-delimiter incorrectly.

\subsubsection{Cluster with Expert guidance}
In this step, we employed a semantic clustering approach which grouped responses based on the underlying themes or sentiments ("semantic clusters") they conveyed.
In preliminary explorations, we found that the LLM is able to automatically generate interesting semantic clusters from a list of the subjects' responses without additional human guidance. 
However, in many cases (21 out of $\approx$40 questions), we felt like it was important to exercise more control over the LLM's response to facilitate human consumption of the results.
So, we provided expert guidance in the form of a list of primary clusters or "themes" (defined on a per-question basis), which were included in the prompt using a template (detailed in \ref{supp_prompt_cluster}).
Secondary clusters or "sub-themes" were discovered automatically by the LLM.
Each subject's response was mapped exclusively to one primary cluster and could furthermore be associated with one or more secondary clusters.

LLM Response Validation: 
The LLM returned two lists in its response: one of the cluster labels and their definitions, and the other of the cluster-labels (single or two-level clustering) assigned to each subject.
The LLM was called once for each of 40 questions, and these responses were parsed using the Python Pandas library. 
A LLM response is marked invalid if it was ill-formatted or incomplete. 
The LLM was called again on invalid responses till the LLM returned a valid response. 
We found that almost 20 questions would have invalid responses in the first attempt, and in total, we were making almost 80\% additional calls to get valid responses for all questions. 
We suspect that the rate of invalid responses in this step is higher than in the previous step due to the added complexity of the task i.e. the response needs to first produce a valid clustering-schema, and then additionally assign each of 93 subjects to the clusters according to the clustering schema.

\subsubsection{Recluster}
We repeated the above clustering step four additional times using a similar prompt (detailed in \ref{supp_prompt_recluster}) with previously defined cluster-definitions.
% We re-ran the clustering process 4 additional times, for a total of 5 clustering trials per question. 
% The same expert-provided or LLM-discovered cluster definitions were used in each trial. 
As in the original clustering, any invalid LLM responses were automatically detected and re-processed until a valid response was obtained.
For the final cluster assignments used in downstream analysis, we applied a majority vote rule based on the 5 clustering repetitions. 
% That is, each subject was assigned to the cluster they most commonly belonged to across the trials. 
% This robustified approach helps account for occasional variability in the LLM outputs.
% Since we compare human cluster assignments with `robust' cluster assignments generated by the LLM, 
In a few cases ($<1\%$ of all subject-question pairs), this process failed to find any cluster assignments that passed the majority-vote. 
% (See \ref{Sup3_confidence}).

% -----------------
\subsection{Human evaluation of LLM responses}
Our study employed human evaluation to verify the alignment between RACER-generated clusters and human interpretation, utilizing two independent evaluators who analyzed the responses of 20 randomly selected subjects from a pool of 93.
Each evaluator individually reviewed the raw interview transcript files for the selected 20 subjects 
% (Subject IDs 3, 4, 7, 14, 18, 19, 28, 33, 34, 35, 36, 41, 42, 48, 56, 63, 84, 99, 100, 102) 
and used the same cluster definitions as RACER to assign subjects to clusters.
% Each evaluator assigned cluster(s) to the responses corresponding to questions Q.14 to Q.41, based on the cluster definitions provided in advance. 
% These clusters were initially defined by expert opinion, i.e, the primary cluster (example: C1,C2, C3,etc.), followed by sub-cluster generation (example: C1.1, C1.2, C2.1, C2.2, etc) by the LLM.  
Human evaluators spent approximately 30 minutes per subject on average for a comprehensive review and categorization of the responses. 
This time investment reflects the thoroughness and attention to detail applied by the evaluators in their analysis, and also highlights the limits of this process to scale to large study populations.  
% Human evaluators required an average of 30 minutes per subject, reflecting the evaluators' detailed and careful approach.
To validate the semantic clustering results produced by the LLM, each human evaluator compared their assigned scores with those generated by the LLM. 
An inter-rater comparison was also conducted, involving a detailed examination of the scores and evaluations independently made by both human evaluators (E1 and E2) for the same set of subjects. 
Concordance scores of 1 were assigned to clusters that precisely matched or were sub- or super-sets of each other, while discrepancies received a concordance score of 0. 
The overall concordance ratio represented the proportion of clusters aligning between the evaluators.

% Furthermore, to establish a comprehensive understanding, these human-generated evaluations were compared with the results obtained from the RACER-assigned clusters. This multifaceted approach aimed to assess not only consistency between human evaluators but also the alignment or divergence between human assessments and those produced by the LLM. The inter-rater comparison, coupled with the comparison to LLM-generated results, serves as a critical step in ensuring the reliability and validity of the entire evaluation process.
Additionally, the evaluators' findings were juxtaposed with RACER's cluster assignments to gauge both inter-evaluator consistency and the degree of correspondence with the LLM's outcomes.
% For each question, the Cohen's kappa coefficient between evaluators correlated with the same pair's concordance score (Appendix \ref{Sup1_cohens}), and showed similar differences between evaluator pairs as concordance. 
We also compared the use of Cohen's kappa coefficient with our concordance score and found them to be similar (Appendix \ref{Sup1_cohens}).
Due to the nature of the comparison across questions which varied in the number of possible clusters as well as probability of different cluster assignment across questions, the concordance scores were used as they better described the intended comparisons.
Instances where RACER did not produce any robust cluster assignments were categorized as 'mismatch' during the evaluation process.

% \subsubsection{Statistical analyses}
% Statistical analyses were performed with Prism 10 (GraphPad Software).
% Chi-square test was used to calculate if proportional distribution of concordance scores or confidence scores significantly differed between conditions and groups. 
% Matched Analysis of variance (ANOVA) or a non-parametric equivalent (Friedman’s test) with multiple comparison test for all pairs of groups was used to determine if there is a statistically significant difference between evaluator pairs.
% Two-way ANOVA was applied for comparing independent variables (questions and evaluator pair, subject and evaluator pairs, or category and evaluator pairs). 
% Anderson–Darling test, D’Agostino–Pearson, Shapiro–Wilk, and Kolmogorov–Smirnov tests were used to determine if data were normally distributed. 
% Spearman’s rank correlation was used to determine correlation. 
% An online Graphpad calculator was used to determine the Cohen’s Kappa coefficients for each question between evaluator pairs.

%%%%%%%%%%%%%%%%%%%%%%%%%%%%% 
\section*{Acknowledgements}
\noindent We thank the involved research staff and consenting subjects for supporting this research.

%%%%%%%%%%%%%%%%%%%%%%%%%%%%% 
\section*{Author Contributions Statement}
% \textcolor{red}{\textbf{
\noindent SHS, NM, AS and AP conceived of the study/analysis. 
NM supervised the collection of interviews.
SHS designed and coded the automated interview transcription, LLM processing, data-analysis, and visualization pipelines.
SHS and NM evaluated early LLM performance and designed the expert-guided clustering criteria.
KJ and KB performed the human evaluation, data interpretation and writing related to human evaluation. 
KJ visualized and analyzed the human evaluation data.
AP funded the transcription and large-language model accounts.
SHS wrote the first draft of the manuscript, excluding text related to human evaluation.
All authors interpreted the results and reviewed and edited the manuscript. 
% }}

%%%%%%%%%%%%%%%%%%%%%%%%%%%%% 
% https://www.nature.com/npjdigitalmed/for-authors-and-referees/submission-guidelines

\section*{Competing Interests Statement}
% \textcolor{red}{\textbf{
\noindent All authors declare no financial or non-financial competing interests.
% }} 

\section*{Code Availability Statement}
\noindent Code used to generate our results can be found at \\
\url{https://github.com/satpreetsingh/RACER}

\section*{Data Availability Statement}
\noindent Aggregated cluster assignment data required to reproduce figures has been made publicly available on the accompanying code repository. 
Human evaluation statistical analyses and visualizations were performed with Prism 10 (GraphPad Software).
Data from earlier stages of the pipeline contain Personally identifiable information (PII) and therefore has not been released.

%%%%%%%%%%%%%%
% \clearpage
% \small
% \newpage
\bibliographystyle{unsrt} 
\bibliography{main}

%%%%%%%%%%%%%%%%%%%%%%%%%%%%%%%%%  Appendices  %%%%%%%%%%%%%%%%%%%%%%%%%%%%%%%%%%%%% 
\clearpage
\onecolumn
\appendix
\noindent \textbf{APPENDICES}

%%%%%%%%%%%%%%%%%%%%%%%%%%%%%%%%%
\section{LLM prompt for retrieving relevant responses from interview transcripts}
\label{supp_prompt_retrieve}

\begin{lstlisting}
Here is a template (tab-separated-values) of an interview (conducted in 2020) between an interviewer and a healthcare professional or medical student. 
Populate the 'answer' column of the template below using the interview transcript appended after the template. 
Be sure to note any positive, negative or neutral emotions expressed by the interviewee in the answer.
If a template question was not asked in the appended transcript (or is not applicable), the answer should be "NA".
For the last 'evidence' column, provide evidence, by quoting verbatim (except for newlines) the parts of the transcript that were most relevant to answering the question.

question_number	question	answer	evidence
1	How old are you?	[numeric]
2	Where do you live?	[city, state, country]
3	What is your marital status?	[single/married/divorced/widowed/etc]
4	Do you have kids?	[yes/no]
5	If you do have kids, provide details	[details]
6	Are you a caretaker otherwise? (if not own kids, eg elderly parents, adopted family member, etc)	[yes/no; details]
7	What type of healthcare professional or student/trainee are you?	[details]
8	If student or trainee, what year are you in?	[year of program]
9	What institution did you complete your (or are currently) training at?	[name and location of institution]
10	If you are a physician, did you train in the US at any point?	[yes/no]
11	What is your specialty (if student, what specialty are you thinking of)?	[details]
12	How long have you been practicing?	[in years, or NA for student]
13	Over the past two months, have you practiced clinically in areas where you could be in touch with patients who have covid-19?	[yes/no]
14	Are you concerned about your safety, and how?	[yes/no; details]
15	Are you concerned about safety of loved ones, and how?	[yes/no; details]
16	Have you modified your routine to protect yourself or others, and how?	[yes/no; details]
17	Has this crisis taken a toll on you physically in any way?	[yes/no; details]
18	How many hours are you working on average (per week) nowadays?	[numeric]
19	How has your working schedule and logistics changed?	[details]
20	How do your working hours compare to pre-covid-19 crisis?	[details]
21	How do think the covid-19 crisis has affected the quality of patient care?	[details]
22	How has it changed your approach to management? (different from usual, at odds with existing guidelines, may not be as effective, etc.)	[details]
23	Are your processes different for end-of-life decisions? Do you have to take people off ventilator more frequently?	[details]
24	How prepared do you feel for the COVID-19 pandemic on a personal level?	[details]
25	How prepared do you feel the unit/hospital is for the COVID-19 pandemic?	[details]
26	How do you think this crisis has affected you emotionally?	[note emotions recognized from interviewee;details]
27	Do you feel supported by peers and/or family during this time?	[details]
28	Have you had more problems with family during this time?	[details]
29	Before this crisis, if someone asked you about your burnout level, what would you have answered?	[score (e.g. 6 out of 10) and/or details]
30	How burned out do you feel nowadays (during the ongoing COVID crisis)?	[score (e.g. 6 out of 10) and/or details]
31	How do you feel about working from home OR at the frontlines?	[Home/Frontlines/Other; details]
32	Do you feel you should be able to handle this as a healthcare professional?	[yes/no; details]
33	What impact do you see this crisis having on you in the near future?	[details]
34	What impact do you see this crisis having on you about five years from now?	[details]
35	Would you seek help if you felt burned out? How?	[yes/no; details]
36	Would you change jobs or career trajectories?	[yes/no; details]
37	Has this crisis affected your specialty decision or career plans in any way?	[yes/no; details]
38	Would you get (professional or other) help/care if you felt mentally overwhelmed? How? When?	[yes/no; details]
39	Any obstacles you foresee in getting help if you needed to?	[yes/no; details]
40	If student or trainee, how closely do you feel that you are adhering to the Hippocratic oath during this time?	[closely/not-closely; details]
41	If student or trainee, do you agree with your school's policies regarding medical students' roles at this time?	[yes/no; details]

TRANSCRIPT:
\end{lstlisting}
\textit{[Interview Transcript Appended]}

%%%%%%%%%%%%%%%%%%%%%%%%%%%%%%%%%
\clearpage
\section{LLM prompt template for semantic Clustering of responses aggregated across all subjects}
\label{supp_prompt_cluster}
\noindent Out of 40 questions in our template in \ref{supp_prompt_retrieve}, 21 questions had expert-provided templates that defined the primary clusters but left secondary-cluster definitions to the LLM. 
4 questions (Q13, Q15, Q17, Q19) used LLM-discovered single-level clustering.
The following Python code shows the template used for generating the prompt associated with each question: 

\begin{lstlisting}
TEMPLATE = """Cluster the responses in the table below at two levels.
Top level clusters must be {clusters}.
Top level clusters have mutually-exclusive cluster membership.
For the next level, cluster the responses from subjects belonging to each top-level cluster highlighting the common theme per cluster.
Subjects can belong to multiple clusters at this level. 

Your response should be in tab-separated-values format, with the following columns:
subject_id  top_level_cluster_id    secondary_cluster_ids

Example output line: 
C-002   C1  "C1.1,C1.2,C1.4"

Start your response by defining each top and secondary cluster in tab-separated-values format, with columns: 
cluster_id  cluster_name    cluster_description

Note that some subject_ids may not be present in the prompt, and so should also not be present in your response.
Provide both the (tab-separated) cluster-definitions table and the (tab-separated) cluster-assignments table in your response.
\n"""

prompts = {
    "default": """Cluster the responses in the table below highlighting the common theme per cluster.
Group subjects that provide unclear, irrelevant, or no responses into a separate "excluded" cluster.
Subjects can belong to multiple clusters. Your response should be in tab-separated-values format, 
with the following columns: subject_id, cluster_ids

Example output line: 
subject_id  cluster_ids
C-002   "C2,C3"

Start your response by defining each cluster in tab-separated-values format, with columns: 
cluster_id, cluster_name, cluster_description

Note that some subject_ids may not be present in the prompt, and so should also not be present in your response.
Provide both the (tab-separated) cluster-definitions table and the (tab-separated) cluster-assignments table in your response.
\n""",
    0: TEMPLATE.format(
        clusters="(1) Young Adults (22 to 33), (2) Middle-aged Adults (34 to 45), (3) Older Adults (46 to 60), (4) Seniors (61 and above), and (5) Unclear/irrelevant/no response"
    ),
    1: TEMPLATE.format(
        clusters="(1) Houston, Texas, (2) San Antonio, Texas, (3) Texas (Other), (4) Florida, (5) Mid-West US, (6) US (Other) and (7) Unclear/Excluded/No response"
    ),
    2: TEMPLATE.format(
        clusters="(1) Not currently married, (2) Married currently, and (3) Unclear/Excluded/No response"
    ),
    14: TEMPLATE.format(
        clusters="(1) Yes, (2) No, and (3) Unclear/irrelevant/no response"
    ),
    16: TEMPLATE.format(
        clusters="(1) Yes, (2) No, and (3) Unclear/irrelevant/no response"
    ),
    # 17: Numeric: How many hours are you working on average (per week)?
    17: TEMPLATE.format(
        clusters="(1) Full-time, (2) Less than Full-time, (3) More than Full-time, and (4) Unclear/Excluded/No response"
    ),
    18: TEMPLATE.format(
        clusters="(1) Increased hours, (2) Decreased hours, (3) No change, (4) Other, and (5) Unclear/irrelevant/no response"
    ),
    # 19: How does this compare to pre-covid-19 crisis?
    19: TEMPLATE.format(
        clusters="(1) Increased hours, (2) Decreased hours, (3) No change, (4) Other, and (5) Unclear/irrelevant/no response"
    ),
    20: TEMPLATE.format(
        clusters="(1) Better, (2) Worse, (3) No-change, (4) Other and (5) Unclear/irrelevant/no response"
    ),
    21: TEMPLATE.format(
        clusters="(1) Changed, (2) No change, (3) Fluctuating/uncertain change, and (4) Unclear/irrelevant/no response"
    ),
    22: TEMPLATE.format(
        clusters="(1) Yes, (2) No, and (3) Unclear/irrelevant/no response"
    ),
    23: TEMPLATE.format(
        clusters="(1) Prepared, (2) Unprepared, and (3) Unclear/irrelevant/no response"
    ),
    24: TEMPLATE.format(
        clusters="(1) Prepared, (2) Unprepared, and (3) Unclear/irrelevant/no response"
    ),
    25: TEMPLATE.format(
        clusters="(1) Positively (e.g. excitement), (2) Negatively, (3) Mix of Positively and Negatively, (4) Neutral, and (5) Unclear/irrelevant/no response"
    ),
    26: TEMPLATE.format(
        clusters="(1) Yes, (2) No, (3) Mixed, (4) Fluctuating over time and (5) Unclear/irrelevant/no response"
    ),
    27: TEMPLATE.format(
        clusters="(1) Yes, (2) No, and (3) Unclear/irrelevant/no response"
    ),
    28: TEMPLATE.format(
        clusters="(1) No/Mild (e.g. 1, 2 or 3 out of 10), (2) Moderate (e.g. 4, 5 or 6 out of 10), (3) Severe (e.g. 7, 8, 9 or 10 out of 10), and (4) Unclear/irrelevant/no response"
    ),
    29: TEMPLATE.format(
        clusters="(1) No/Mild (e.g. 1, 2 or 3 out of 10), (2) Moderate (e.g. 4, 5 or 6 out of 10), (3) Severe (e.g. 7, 8, 9 or 10 out of 10), and (4) Unclear/irrelevant/no response"
    ),
    30: TEMPLATE.format(
        clusters="(1) Positively (e.g. excitement), (2) Negatively, (3) Neutral/Mixed and (4) Unclear/irrelevant/no response"
    ),
    31: TEMPLATE.format(
        clusters="(1) Yes, (2) No, (3) Mixed, and (4) Unclear/irrelevant/no response"
    ),
    32: TEMPLATE.format(
        clusters="(1) Positive, (2) Negative, (3) Neutral/Mixed and (4) Unclear/irrelevant/no response"
    ),
    33: TEMPLATE.format(
        clusters="(1) Positive, (2) Negative, (3) Neutral/Mixed and (4) Unclear/irrelevant/no response"
    ),
    34: TEMPLATE.format(
        clusters="(1) Yes, (2) No, and (3) Unclear/irrelevant/no response"
    ),
    35: TEMPLATE.format(
        clusters="(1) Yes, (2) No, and (3) Unclear/irrelevant/no response"
    ),
    36: TEMPLATE.format(
        clusters="(1) Yes, (2) No, and (3) Unclear/irrelevant/no response"
    ),
    37: TEMPLATE.format(
        clusters="(1) Yes will get professional help, (1) Yes but not professional help, (3) Mixed, (4) Will not seek/get help and (5) Unclear/irrelevant/no response"
    ),
    38: TEMPLATE.format(
        clusters="(1) Yes, (2) No, and (3) Unclear/irrelevant/no response"
    ),
    39: TEMPLATE.format(
        clusters="(1) Adhering Closely, (2) Not adhering closely OR Adhering conditionally, and (3) Unclear/irrelevant/no response"
    ),
    40: TEMPLATE.format(
        clusters="(1) Yes, (2) No, (3) Mixed/Conditionally, and (3) Unclear/irrelevant/no response"
    ),
}
\end{lstlisting}

%%%%%%%%%%%%%%%%%%%%%%%%%%%%%%%%%
\clearpage
\section{LLM prompt for Re-Clustering using previously defined clusters}
\label{supp_prompt_recluster}

\begin{lstlisting}
Cluster the responses in the table below highlighting the common theme per cluster.
Group subjects that provide unclear, irrelevant, or no responses into a separate "excluded" cluster.
Subjects can belong to multiple clusters. Your response should be in tab-separated-values format, 
with the following columns: subject_id, cluster_ids

Example output line: 
subject_id  cluster_ids
C-002   "C2,C3"

Note that some subject_ids may not be present in the prompt, and so should also not be present in your response.
Provide both the (tab-separated) cluster-definitions table and the (tab-separated) cluster-assignments table in your response.

subject_id	Are you a caretaker otherwise? (if not own kids, eg elderly parents, adopted family member, etc)
C001	No
C002	No
C003	No
C004	No
C005	No

...

C086	Yes, looks after his mother-in-law's finances
C087	No
C090	Yes; Partial caretaker for parents
C099	No
C100	No
C101	No
C102	No

Use the following cluster definitions (Do not repeat this in output):
cluster_id	cluster_name	cluster_description
C1	Caretakers of Family Members	Subjects who responded that they take care of relatives (elderly parents, children, siblings or others).
C2	Caretakers of Animals	Subjects who take care of animals.
C3	Partial Caretakers	Subjects who participate in caretaking but not as primary caretakers.
C4	Financially Supportive	Subjects who provide financial support instead of physical caretaking.
C5	No Caretaking Responsibilities	Subjects who stated that they do not take care of anyone.
C6	Excluded	Responses that are unclear, irrelevant, or did not provide a response to the question.
\end{lstlisting}

%%%%%%%%%%%%%%%%%%%%%%%%%%%%% #############################
% \input{supplement}
\clearpage

\section{Concordance analysis}
\label{supp_concordance}

\begin{figure*}[tbhp!]
\centering
\includegraphics[width=0.85\linewidth]{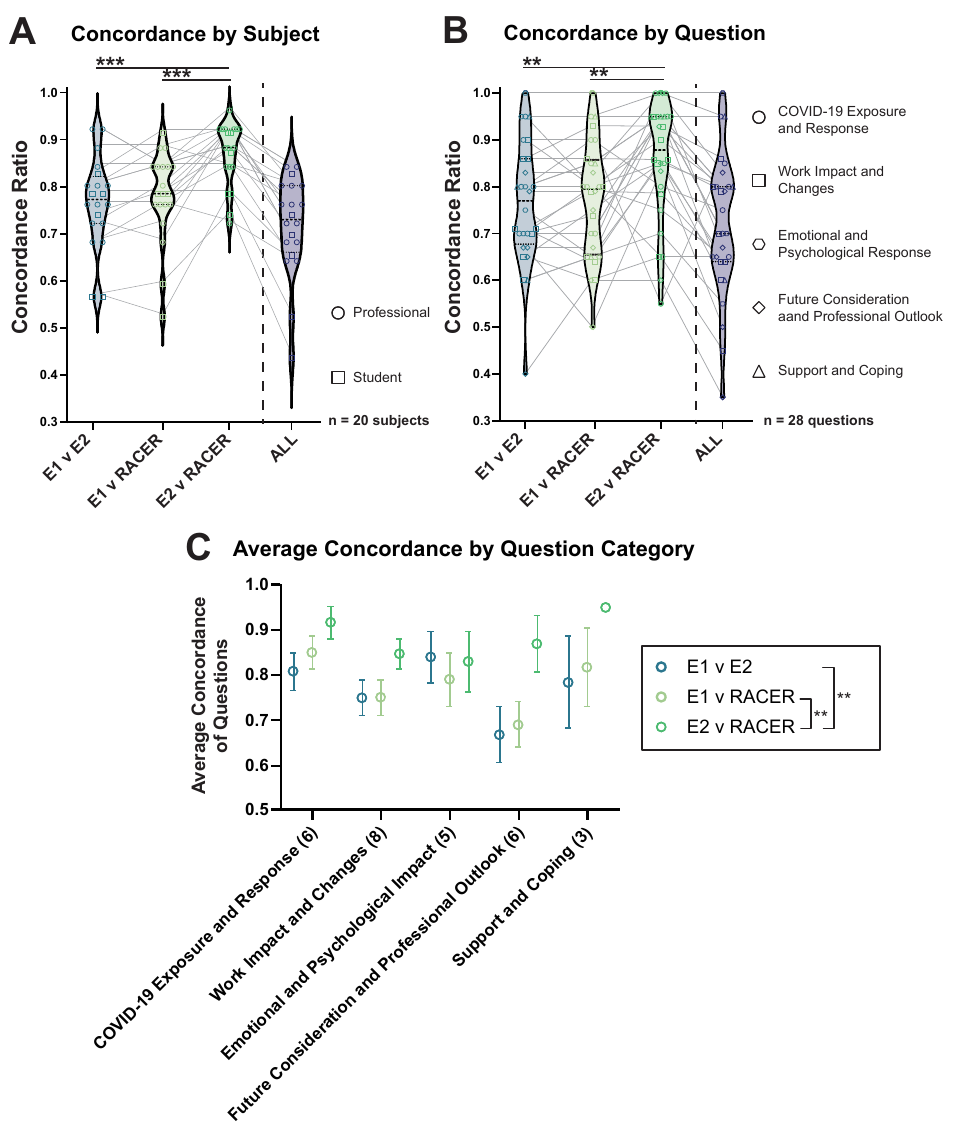}
% \includesvg[width=0.80\linewidth]{appendix_b_racer.svg}
\caption{
%\centering
\textbf{Differences in subject and question concordance ratio accounts for most inter-evaluator variability :} (A) Average concordance score of subjects compared across evaluator pairs. Average subject concordance compared between evaluator pairs (E1 v E2, E1 v GPT, and E2 v GPT) using 2Way ANOVA found a that the subject accounts for 57.4\% of the variability (p $<$ 0.0001) while differences between evaluator pairs account for 22.5\% of the variability (p $<$ 0.0001). Post-hoc Tukey’s multiple comparison across evaluator pairs found that the subject concordance of E2 v GPT was significantly different than both E1 v GPT and E1 v E2. (B) Average concordance score of questions compared across evaluator pairs. Average question concordance compared between evaluator pairs using 2Way ANOVA found a that the question accounts for 58.1\% of the variability (p $<$ 0.0001) while differences between evaluator pairs account for 9.8\% of the variability (p$<$0.0001). Post-hoc Tukey’s multiple comparison across evaluator pairs found that the subject concordance of E2 v GPT was significantly different than both E1 v GPT and E1 v E2. (C) Concordance score of questions were averaged across categories (mean +/- SEM) and compared across evaluator pairs. The average category concordance compared between evaluator pairs using 2Way ANOVA found that the category accounts only for 10.8\% of the variability (p = 0.032) while differences between evaluator pairs account for 14.6 \% of the variability (p = 0.001). Post-hoc Tukey’s multiple comparison across evaluator pairs found that the subject concordance of E2 v GPT was significantly different than both E1 v GPT and E1 v E2. * p $<$ 0.5, ** p $<$ 0.01, *** p $<$ 0.001, **** p $<$ 0.0001. 
%}
}
\label{fig2_humanval_supp}
\end{figure*}

\newpage
\section{Concordance vs Cohen's Kappa}
\label{supp_concordance_vs_cohen}

\begin{figure*}[tbhp!]
\centering
\includegraphics[width=0.95\linewidth]{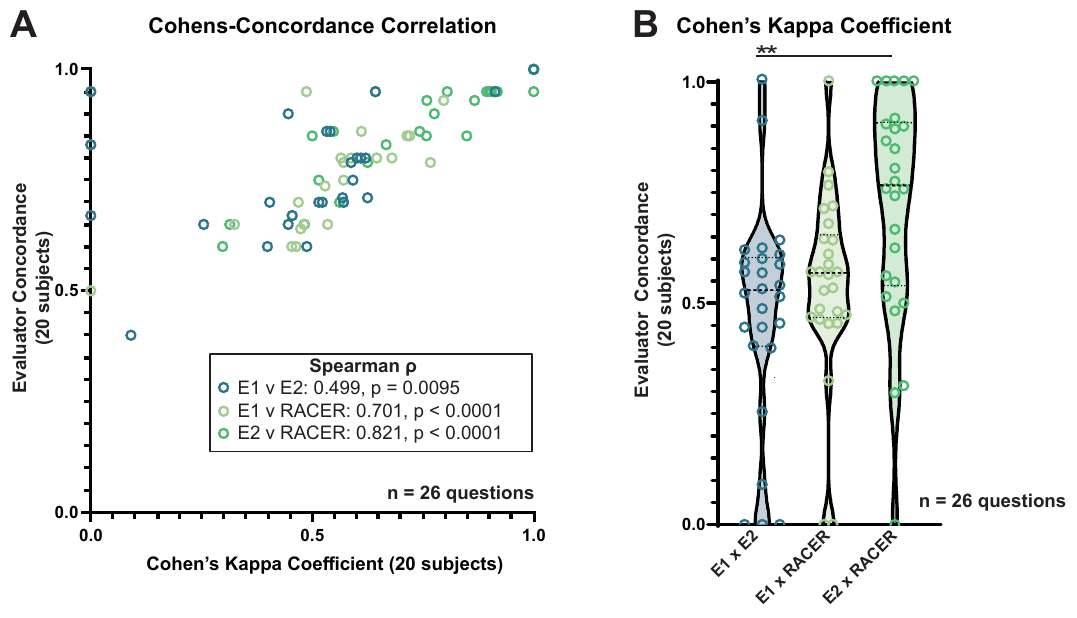}
% \includesvg[width=0.95\linewidth]{appendix_a_racer.svg}
\caption{
%\centering
\textbf{Cohen’s kappa coefficient correlates with the concordance ratio}: (A) The Cohen’s Kappa coefficient calculated for each question correlates significantly with the evaluator concordance score using Spearman rank correlation. (B) The distribution of Cohen’s Kappa coefficient for each question between different evaluator pairs differs (Friedman test, p = 0.0015 with Dunn’s multiple comparison). * p $<$ 0.5, ***  p $<$ 0.01, *** p $<$ 0.001, **** p $<$0.0001.
%}
}
\label{Sup1_cohens}
\end{figure*}

\end{document}